\pdfoutput=1

\documentclass[11pt]{article}

\usepackage[]{acl}

\usepackage{times}
\usepackage{latexsym}

\usepackage[T5]{fontenc}

\usepackage[utf8]{inputenc}

\usepackage{microtype}
\usepackage{graphicx}
\usepackage{threeparttable}
\usepackage{multirow}
\usepackage{import}
\usepackage{svg}

\title{ViWOZ: A Multi-Domain Task-Oriented Dialogue Systems Dataset For Low-resource Language}

\author{Author 1 \\ Address line \\  ... \\ Address line
        \AND
        Author 2 \\ Address line \\ ... \\ Address line \And
        Author 3 \\ Address line \\ ... \\ Address line}

\author{Phi Nguyen Van \\
  \texttt{phinv@vnu.edu.vn} \\\And
  Tung Cao Hoang \\
  \texttt{19020055@vnu.edu.vn} \\\And
  Dung Nguyen Manh \\
  \texttt{18020370@vnu.edu.vn} \\\AND
  Quan Nguyen Minh \\
  \texttt{19020019@vnu.edu.vn} \\\And
  Long Tran Quoc \\
  \texttt{tqlong@vnu.edu.vn} \\}

\begin{document}
\maketitle

\begin{abstract}
Most of the current task-oriented dialogue systems (ToD), despite having interesting results, are designed for a handful of languages like Chinese and English. Therefore, their performance in low-resource languages is still a significant problem due to the absence of a standard dataset and evaluation policy. To address this problem, we proposed \textbf{ViWOZ}, a fully-annotated Vietnamese task-oriented dialogue dataset. \textbf{ViWOZ} is the first multi-turn, multi-domain tasked oriented dataset in Vietnamese, a low-resource language. The dataset consists of a total of 5,000 dialogues, including 60,946 fully annotated utterances. Furthermore, we provide a comprehensive benchmark of both modular and end-to-end models in low-resource language scenarios. With those characteristics, the \textbf{ViWOZ} dataset enables future studies on creating a multilingual task-oriented dialogue system.
\end{abstract}

\section{Introduction}

Task-oriented dialogue systems (ToD), a narrowed version of a general dialogue system, allow users to interact with virtual agents to accomplish certain tasks. In recent years, task-oriented has become an active research topic thanks to the development of neural network models. \cite{ren-etal-2018-towards,wu-etal-2020-tod,henderson-etal-2020-convert}. Despite the accelerating movement of performances, those works are limited to hands-on languages like English. \cite{budzianowski-etal-2018-multiwoz} and Chinese \cite{zhu-etal-2020-crosswoz} due to the lack of large-scale multilingual languages. 

In the field of task-oriented dialogue systems, Natural Language Understanding (NLU) is the only module having diverse language datasets. MultiATIS++ \cite{xu-etal-2020-end} is a dataset translated from ATIS \cite{price-1990-evaluation}, which covers 9 languages but only has a single domain of airline travel, and most of the languages are also from the same language family. Along with MultiATIS++, several researchers have translated the ATIS into individual languages like Chinese \cite{6639292}, Vietnamese \cite{nguyen-nguyen-2021-phonlp}, and many other languages \cite{susanto-lu-2017-neural,8461905,xu-etal-2020-end}. Extending the dataset into multi-domain, MTOP \cite{li-etal-2021-mtop} and SID \cite{van-der-goot-etal-2021-effectiveness} are both multi-domain and multi-lingual dialogue datasets, but those datasets are still limited to NLU annotations only. Despite having multilingual diversity, simply translating and mapping the dataset will make the dialogue become tedious, lacking locale, and trivial, leading to over-optimistic results. \cite{ponti-etal-2020-xcopa, artetxe-etal-2020-translation}. 

For multi-lingual Dialogue State Tracking (DST), \cite{mrksic-etal-2017-semantic} translated the WoZ 2.0 DST dataset \cite{wen-etal-2017-network} into German and Italian. DSTC 5 and DSTC 6 \cite{kim2016fifth, hori2019overview} show the benchmark of DST models on zero-shot cross-lingual transfer from English to Chinese and transferring dialogue knowledge from English to Japanese. Finally, DSTC 9 \cite{gunasekara2020overview} introduces the first challenge to benchmark cross-lingual DST systems on large scale datasets, focused on transfer between English and Chinese, using MultiWOZ 2.1 \cite{eric-etal-2020-multiwoz} as the English dataset and CrossWOZ \cite{zhu-etal-2020-crosswoz} as the Chinese dataset. Those datasets enable various methods in both modular and end-to-end ToD.

As far as we know, there is still no large-scale, multi-domain dataset for task-oriented dialogue systems in Vietnamese, a low-resource language scenario. Therefore, we proposed \textbf{ViWOZ}, built upon MultiWOZ 2.1, as the first fully annotated multi-domain Vietnamese task-oriented dialogue dataset. Our dataset has the following advantages compared to previous works:

\begin{table*}
\centering
\begin{tabular}{l|ccccc}
 &  MultiWOZ  & RiSAWOZ & CrossWoz  & BiToD & ViWOZ\\
\hline 
Language(s)  &  EN & ZH & ZH  &  EN, ZH & \textbf{VI}\\
No. Dialogues & 8,438 & 10,000 & 5,012 & 5,787 & \textbf{4014}\\
No. domains & 7 & 12 & 5 & 5 &  \textbf{7} \\
 No. turns & 115,434 & 134,580 & 84,692 & 115,638 & \textbf{48,944}\\
 Avg turns & 13.46 & 13.5 & 16.9 & 19.98 & \textbf{12.19}\\
 Slots & 25 & 159 & 72 & 68 & \textbf{25}\\
Values & 4,510 & 4,061 & 7,871 & 8,206 & \textbf{4,510}\\
\end{tabular}
\caption{\label{Table 1}
Comparisons of training set with other ToD datasets.
}
\end{table*}

\begin{itemize}
  \item \textbf{ViWOZ} is the first fully annotated multi-domain task-oriented dialogue dataset in Vietnamese, a low-resource language.
  \item In previous ToD datasets, researchers simply translated the original corpora into the target language, which could make the dataset unrepresentative of realistic conversation due to a lack of culture and locale. Furthermore, existing Machine Translation system such as Google Translate has known issues of making translation errors. In our approach, diverse paraphrases from crowd-sourced workers resolve the translations errors and increase the naturalness of conversations. It is the first fully annotated multi-domain task-oriented dialogue dataset in Vietnamese, a low-resource language.
  \item \textbf{ViWOZ} is built upon MultiWOZ 2.3, providing fully annotated dialogue states and dialogue acts for both the system side and user side. We also corrected the errors in the original data when mapping from English to Vietnamese.
  \item The effectiveness of multilingual/cross-lingual/monolingual language models on low resource languages, as well as the performance of current approaches, remains an open question. \cite{razumovskaia2021crossing}. In this work, we conduct intensive experiments in different settings as an initial step to uncover the multilingual ToD problem.
\end{itemize}

\section{Related Work}

According to \cite{budzianowski-etal-2018-multiwoz}, there are three categories of task-oriented dialogue datasets: human-to-human, human-to-machine and machine-to-machine. Each category of datasets is associated with whether the user and system agent are machine or human. Regardless of dataset category, most of them are single-domain or only available in high-resource languages or both. As far as our limited knowledge, there is still no large-scale multi-domain dataset for low-resource languages. 

\textbf{Human-to-Human} is a type of dataset where both the user and the system are human agents. The crowd-sourced workers are hired to talk to each other and are given some instructions. This setup is widely used in both single-domain \cite{kim2016fifth} and large-scale multi-domain scenarios \cite{eric-etal-2020-multiwoz}. The dataset built upon this setup creates a diverse and realistic dialogue, but it also requires intensive human effort in data creation and quality control.

\textbf{Human-to-Machine} is a type of dataset where humans interact with a dialogue system. 
This setup is already used to create the dataset \cite{hori2019overview} from The Dialogue State Tracking
Challenges (DSTC). Despite the automation on the system side, the quality of those datasets is heavily influenced by the quality of the dialogue system.

\textbf{Machine-to-Machine} is a type of dataset where both the user and the system are autonomous agents. This fully automated setup enables the creation of large scale datasets with minimal human effort \cite{shah2018building, rastogi-etal-2019-scaling}. However, these datasets still lack lingual diversity.

\textbf{Multilingual ToD dataset} Despite the efforts of several researchers \cite{mrksic-etal-2017-semantic,zhu-etal-2020-crosswoz,lin2021bitod}, one of the major barriers to the widespread application of ToD research \cite{razumovskaia2021crossing} is a lack of multilingual ToD datasets  \cite{razumovskaia2021crossing}. Most of the multilingual datasets only focus on single-domain or only on a handful of languages like English and Chinese. 

\begin{table*}
\centering
\begin{tabular}{lcccccc}
\hline 
 & Train & Dev & Test & Single-domain & Multi-domain & All \\
\hline 
Dialogues        & 4,014       & 493     & 493      & 2,104   & 2,896   & 5000 \\
Turns            & 48,944      & 5,936   & 6,066    & 17,160  & 43,786  & 60,946 \\
Tokens           & 897,550     & 109,453 & 110,897  & 286,684 & 831,214 & 1,117,900 \\
Vocabulary       & 8,994       & 3,072   & 2,908    & 4,563  & 8,538   & 10,136 \\
Avg. turns       & 12.19       & 12.04   & 12.30    & 8.15   & 15.12   & 12.19 \\
Avg. acts        & 1.24        & 1.23    & 1.23     & 1.83    & 1.26    & 1.28 \\
Avg. user-acts   & 1.52        & 1.52    & 1.51     & 1.39    & 1.57    & 1.64 \\ 
Avg. system-acts & 1.64        & 1.65    & 1.63     & 1.58    & 1.66    & 1.52 \\

\hline
\end{tabular}
\caption{\label{Table 2}
\textbf{ViWOZ} dataset statistic. White space tokenization is used to count number of token and vocabulary.
}
\end{table*}

\begin{figure}
\includegraphics[width=8cm]{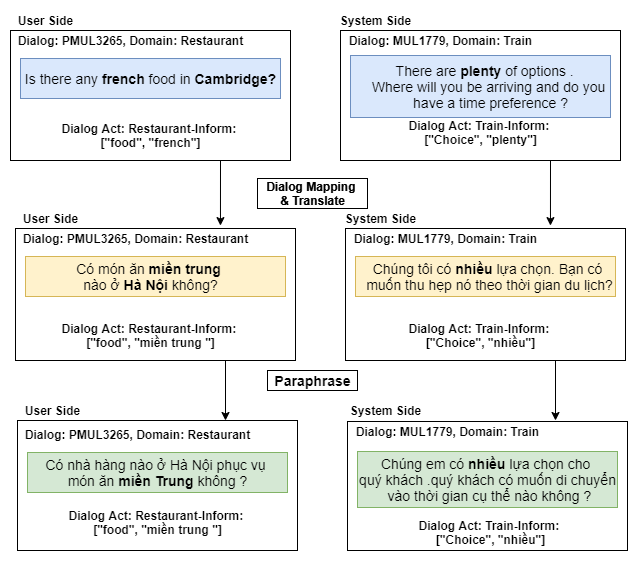}
\caption{\label{Figure 1} Illustration of data construction for user side (dialogue PMUL3265) and system side (dialogue MUL1779)}
\end{figure}

\section{ViWOZ Dataset}

\subsection{Language analysis}
Linguists have conducted researches about the difference of Vietnamese and English. The majority of linguistics researchers consider Vietnamese a Austroasiatic language \cite{language_vietnamese_overview}, while 

\begin{figure}
\includegraphics[width=8cm]{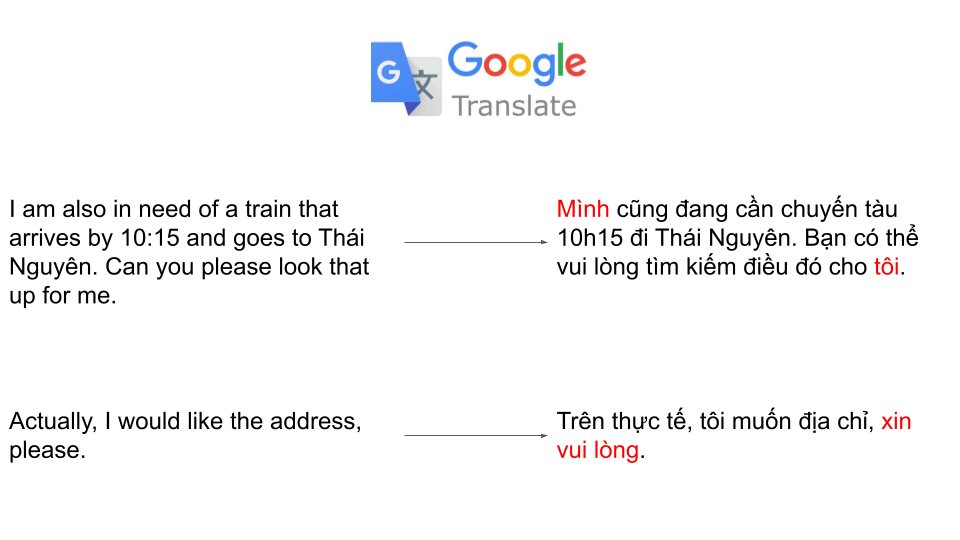}
\caption{\label{Figure 3} Translation of 2 utterances from dialogue PMUL1043 in MultiWOZ 2.3}
\end{figure}

\cite{language_vietnamese_vs_english} has identified a variety of differences between Vietnamese and English, such as lexical, semantic, and grammatical characteristics. For example, when describing an object, a Vietnamese would use the word order $N + Adj$, while an English speaker would use $Adj + N$. Another example is the diversity of pronouns in Vietnamese: a single pronoun \textit{I} representing the speaker could be translated to \textit{Tôi, tao, mình, tớ, anh, chị, em, cô, dì, chú, bác, etc.}, each has a different level of formality, and some could only be used in certain contexts where the speaker knows exactly who he/she is having a conversation with. These linguistic differences, together with the cultural and language usage variations, could be a significant problem for machine translation systems, as grammar or appropriate language usage may not be fully reflected. Figure \ref{Figure 3} demonstrates this problem: In the first example, different mentions referring to \textit{"I"} were used, which is very uncommon in Vietnamese; in the second example, the translation of the adverb \textit{"please"} remains at the end of the utterance, while it is more common to have it in the beginning, i.e \textit{"Xin vui lòng cho tôi địa chỉ"}

\subsection{ViWOZ}
Our dataset is built upon MultiWOZ 2.3 \cite{han2020multiwoz} by translating the dialogues from English to Vietnamese. We also performed manual post-processing to modify the translated data for translation accuracy. A Vietnamese-localized database is also constructed together with the dataset for cultural appropriation. The process of data creation is described in this section.

\subsubsection{Ontology construction}
Task-oriented dialogue systems require a source of truth, also known as an \textit{ontology}, to provide information to the users. MultiWOZ 2.3's ontology consists of a total of 5,946 entities spanning 6 domains (attraction, bus, hospital, hotel, police, restaurant, train). For exact location domain where each entity corresponds to a location in real life, we gathered the same list of locations in Hanoi by crawling from several local online booking and travel agent websites (i.e. traveloka, booking.com, etc.). For the remaining train and bus domains where each item contains a travel route information from a city in England to the main city of Cambridge, we manually map each city in the source data to a real province in Vietnam. Cambridge is mapped to Hanoi, the capital city of Vietnam. All the available values for each item in the ontology are translated to Vietnamese properly. For example, \textit{duration: 51 minutes} is converted into \textit{duration: 51 phút}. The result of this process is a localized ontology in Vietnamese where each entity is uniquely mapped from English.

\begin{figure}
\includegraphics[width=8cm]{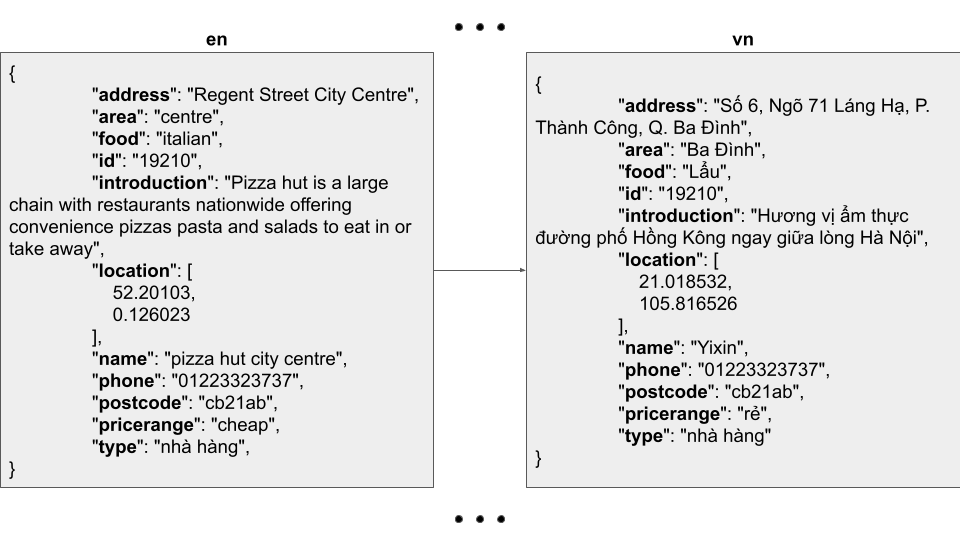}
\caption{\label{Figure 4} Example of a mapping in the $restaurant$ domain of the constructed ontology}
\end{figure}

\subsubsection{Slot Mapping}
Previous datasets like MultiATIS++ simply kept entities intact when translating datasets. This will lead to unnatural dialogues that are unrepresentative of real-life conversations in Vietnamese. In our approach, we use the constructed localized ontology to perform slot mapping based on the utterances and the annotated dialogue acts. Since multiple dialogue act values can represent the same value in the ontology (i.e.$center$ and $centre$ could both mean the center area of Cambridge), we build a dictionary to map annotated dialogue slots from the original values to new values in Vietnamese. 

Manually selecting tens of thousands of values to be mapped in the dictionary can be very time-consuming and inefficient. Therefore, we first build the mapping dictionary by performing a full-text search on the ontology slots. For example, the value in $Restaurant-Name-Inform$ slot of an utterance will be searched in every $name$ slots in the ontology's $restaurant$ domain, the entity with the lowest Levenshtein distance will be selected, and the $name$ slot of the corresponding Vietnamese entity is considered the result of the correct mapping value. The dictionary is then cross-checked and modified manually by two researchers to mitigate the problems of the automated searching algorithm.

Using the generated dictionary, we replaced all the annotated dialogue slots in the utterance and in the annotation with their new value in Vietnamese. The resultant dialogues are in English, with the annotated spans being replaced by the localized entity from the previously constructed dictionary. We apply the mapping process to 5,000 dialogues, randomly sampled from MultiWOZ 2.3. In summary, we have mapped and checked a total of 75,852 entities, 60,946 dialogue acts, and 30,475 dialogue states.

\subsubsection{Dialogue translation and paraphrasing} 
We utilized the Google Translate API to translate every dialogue, with the entities having been replaced by localized ones, from English to Vietnamese. Twelve native speakers are then selected to paraphrase all the translated utterances. Given a translated utterance, the annotators must rewrite the sentence so that the information about the dialogue acts is kept and the language is as natural as possible. Because the original translated value may not be compatible with the paraphrased utterance, the crowd-sourced workers are also required to re-annotate every slot value to rebuild the dialogue state. This ensures that the dialogue is written naturally, as a simple translation of the dataset creates over-simplistic and unrealistic dialogues. The examples in Figure \ref{Figure 1} demonstrate how the paraphrasing process improves the naturalness of the dialogue. The final dataset consists of mapped slots, states, dialogue acts, dialogue goals, and paraphrased utterances for both system and user utterances. Table \ref{Table 2} shows the basic statistics of \textbf{ViWOZ} dataset.

\subsubsection{Human Evaluation}
Multiwoz is a noisy dataset \cite{zang2020multiwoz, ye2021multiwoz}, and after many attempts to rectify the dialogue errors, we still find errors from the original dataset while converting the dialogues from English to Vietnamese. Missing slot value and incorrectly spanned annotation are the two most common errors, which are corrected throughout the annotation of crowd-sourced workers.

Finally, two researchers perform quality assurance by manually reviewing the paraphrased utterances to identify possible syntactic, semantic, and slot annotation errors

\begin{figure}
\includegraphics[width=8cm]{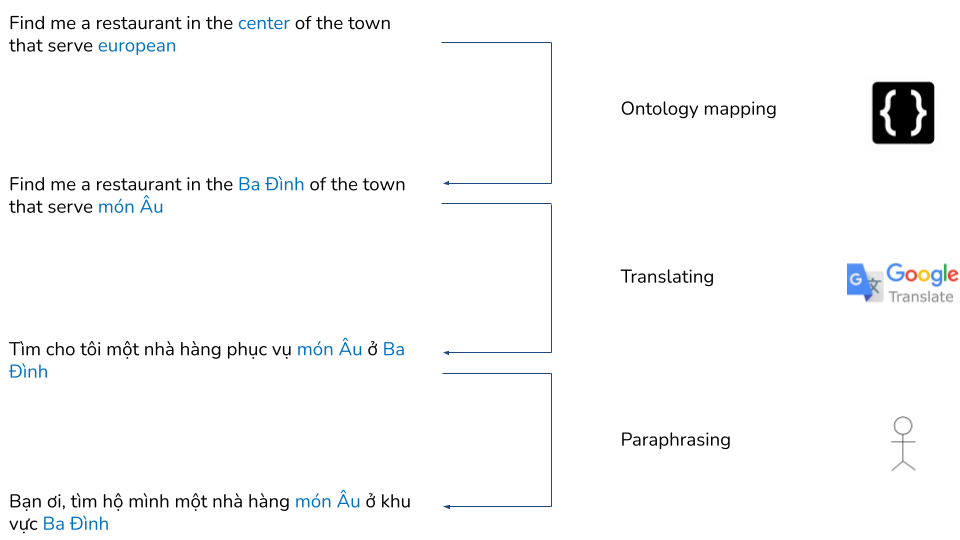}
\caption{\label{Figure 2} The ontology improves dialogue localization, while the paraphrasing acts as human post-processing to improve dialogue naturalness}
\end{figure}

\section{Benchmark and Analysis}

\begin{table*}
\begin{tabular}{llllcccc}
\hline 
\textbf{Task} & \textbf{Model} & \textbf{Pretrained} & \textbf{Metrics} & \textbf{en} & \textbf{zero shot} & \textbf{vn} & \textbf{vn+en}  \\
\hline
\multirow[t]{3}{*}{NLU} & Joint model & PhoBERT & \multirow[t]{3}{*}{Dialogue act F1} &  -  &  -  &  84.70  &  -   \\
& & EnViBERT &  &           88.79 &            44.02 &  \textbf{85.04}  &  \textbf{85.49}   \\
& & XLM-base &  &  \textbf{89.50} &  \textbf{45.89}  &  84.14           &  84.46            \\
\hline
\multirow[t]{3}{*}{DST}& TRADE &   & \multirow[t]{3}{*}{Joint goal accuracy} &  48.62 &  - &  46.61 &  44.63   \\
& SUMBT & PhoBERT  &  &  -    &  -             &  52.95           &  -   \\
&       & EnViBERT &  & 61.86* & 6.17           &  \textbf{54.70}  &  52.09   \\
&       & XLM-base &  & 61.86* &  \textbf{6.20} &  52.00           &  \textbf{52.98}   \\
\hline
\multirow[t]{8}{*}{End-to-end}& MinTL & mT5-small & Inform &  80.04 &  35.5 &  38.1  &  \textbf{43.2}   \\
&    &         & Success &  72.71  &  \textbf{21.3} &  17.8           &  20.3   \\
&    &         & BLEU    &  19.11  & 0.0            &  \textbf{18.1}  &  16.5    \\
&    & mT5-base& Inform  &  82.15  &  34.7          &  46.9           &  \textbf{51.7}   \\
&    &         & Success &  74.44  &  18.1          &  \textbf{35.7}  &  32.7   \\
&    &         & BLEU    &  18.59  & 0.0            &  17.1           &  \textbf{17.2}    \\
\hline

\end{tabular}
    
\caption{\label{Table 3}
Benchmark of a few recent methods on NLU, DST and end-to-end. $-$ cells indicate experiments are not conducted because models do not have the vocabulary of the language. *: result trained on BERT-base from \cite{ye2021multiwoz}.
}

\end{table*}

In ToD, there are two main approaches: modular and end-to-end. While the modular approach aims to create ToD by breaking down the system into specialized modules like Natural Language Understanding (NLU), Dialogue State Tracking (DST), Dialogue Policy, and Natural Language Generation (NLG), end-to-end models often use large language models to directly map the dialogue history to output utterance. With \textbf{ViWOZ}, we can do multiple assessments of those approaches in low resource scenarios. Our main concern is whether current methods, which are optimized for high resource languages like English and Chinese, still work on low-resource language scenarios, especially on the new \textbf{ViWOZ} dataset. We provided benchmarks on various models and settings for two sub-tasks of modular ToD systems: NLU and DST. In addition, we also conduct experiments on end-to-end models.
\begin{itemize}
    \item Performance difference between monolingual, bilingual, and multilingual pre-trained models. This helps us to understand the effectiveness of the monolingual language model in low-resource language ToD.

    \item Effectiveness of model size to performance: whether bigger language models has any advantage on low-resource language ToD.

    \item Zero-shot/bilingual training from cross-lingual pre-trained models: whether high-resource language datasets help boost the performance of low resource ToD.
\end{itemize}

We use the following pre-trained models to conduct the experiments: (1) PhoBERT \cite{nguyen-tuan-nguyen-2020-phobert} is a monolingual language model trained only on Vietnamese corpus, (2) EnViBERT\cite{nguyen20d_interspeech} is a bilingual RoBERTa model train on English and Vietnamese, (3) XLM-RoBERTa \cite{conneau-etal-2020-unsupervised} is a multilingual model trained on 100 languages. 

For the dataset, we use three settings:
\begin{itemize}
    \item \textbf{Zero shot}. To see the effectiveness of multilingual LM in a zero-shot setting, we train the models on the original English dataset and evaluate them on a Vietnamese test set.

    \item \textbf{Monolingual - VN/EN}. In this setting, we train and evaluate models only on Vietnamese data. The EN column is added to Table \ref{Table 3} for convenient comparison between the performance of models in English and Vietnamese.

    \item \textbf{Bilingual - VN+EN}. To investigate knowledge transfer from a high-resource to a low-resource language, we combine the English and Vietnamese training sets into a single training set and test the models on a Vietnamese test set.
\end{itemize}

\subsection{Natural Language Understanding}

NLU is the first module in modular ToD systems. The main purpose of the module is to identify the intent and extract information from user and system utterances. It often consists of two sub-tasks: intent classification and slot filling. There are two main approaches for this module: joint-model and separate model, where the joint model tries to solve two sub-tasks at once, while the separate model solves each task with individual models. Joint-models often have the potential to create a more compact model, and multi-task training objectives also offer better performance \cite{razumovskaia2021crossing}. In cross-lingual NLU, the lack of multilingual data makes zero-shot or few-shot learning the default approaches. By utilizing the massively multilingual Transformers models, the zero-shot model shows strong performance on multilingual NLU datasets. \cite{zhang2019joint, krishnan-etal-2021-multilingual}.

\textbf{Experiment settings}: In this section, we extend BERT NLU \cite{devlin-etal-2019-bert} by replacing BERT with other alternative pre-trained language models. All model configurations (such as the classification head) are the same throughout the experiments, while the language models act as the utterance encoders in a sequence-tagging and multi-label classification joint-training task. By comparing different language models in different training schemes, we can show the effectiveness of pre-trained language models in Natural Language Understanding. We conduct experiments with PhoBERT \cite{nguyen-tuan-nguyen-2020-phobert}, EnViBERT \cite{nguyen20d_interspeech} and XLM-RoBERTa \cite{conneau-etal-2020-unsupervised} for monolingual, bilingual and zero-shot settings, respectively.

\textbf{Result analysis}: The result in the first portion of Table \ref{Table 3} indicates that the bilingual and multilingual models yield a noticeable zero-shot dialogue act F1 score. There is a gap in the monolingual scheme result between \textbf{ViWOZ} and MultiWOZ, since \textbf{ViWOZ} has about half the training samples in comparison to MultiWOZ. In addition, there is little to no difference in the results when training in monolingual data (where the model is trained and validated on Vietnamese only) and in bilingual data (where the model is trained with data from both languages and evaluated on a Vietnamese test set). When taking the model size into account, despite having only about half the number of parameters as XLM-RoBERTa (base), EnViBERT and PhoBERT still achieve competitive results, if not better, suggesting that a well-trained monolingual or bilingual language model could outweigh a larger multi-lingual one.

\subsection{Dialogue State Tracking}

Dialogue State Tracking (DST) takes responsibility for maintaining a dialogue belief state and contains information about the dialogue throughout the conversations \cite{mrksic-etal-2017-neural}. Traditionally, DST often takes implicit input from NLU and previous belief states to produce an updated state of the current utterance. Recently, there has been a trend to utilize long-distance representation of Transformers models to represent and track the whole dialogue.

\textbf{Experiment setting} TRADE (Transferable Dialogue State Generator) \cite{wu-etal-2019-transferable} and SUMBT (Slot-Utterance Matching for Universal and Scalable Belief Tracking) \cite{lee-etal-2019-sumbt} are the two models we used to perform experiments in this part. TRADE generates conversation state using an encoder-decoder design, whereas SUMBT tracks dialogue state using contextual representation from a pre-trained language model. We can demonstrate the function of a pre-trained language model in tracking conversation status by comparing two models with different dataset configurations.

\textbf{Result analysis} The second portion of Table \ref{Table 3}, shows that SUMBT outperforms TRADE across all datasets, indicating the effectiveness of pre-training LM. Mixed Vietnamese and English data hindered model performance, indicating that a mix-language training method produces noisy data and makes it more difficult for models to monitor crucial information while maintaining multilinguality. 

\subsection{End-to-End}

Modular ToD trains and combines each model individually. This pipeline approach leads to error cascading when the performance of later modules depends largely on the performance of previous modules. End-to-end models try to solve this problem by combining modules into a single LM to generate a system response directly from user utterances.

\textbf{Experiment setting} The state-of-the-art end-to-end task-oriented dialogue models MinTL \cite{lin-etal-2020-mintl} are used. We just change the pre-trained LM from T5 \cite{2020t5} to mT5 \cite{xue-etal-2021-mt5} and leave the rest of the implementation unchanged. From MinTL's official code, the hyper-parameters are likewise set to default. All models are trained on a single NVIDIA A6000.

\textbf{Result analysis}: The MinTL failed on all dataset experiment settings; the inform and success scores are significantly lower than the English-only setup, although the BLEU score is similar; the model appears to create reasonable responses but fails to convey accurate information. Furthermore, BLEU scores are similar in monolingual and bilingual setups, showing that there is no improvement when high resource language data is included; this can be explained by the language family difference between English and Vietnamese. The zero shot option, on the other hand, produced results that were equivalent to the monolingual data setting in terms of inform and success scores, demonstrating that multilingual pre-training is beneficial when transferring dialogue structure. When we use mT5-base to increase the model size, we observe that performance on providing correct information improves dramatically from 38.1\% to 46.9\% on monolingual setup and from 43.2\% to 51.7\% on bilingual setup, indicating that larger LMs tend to perform better on downstream tasks.

\section{Conclusion} 
This study presents ViWOZ, the first multi-domain Vietnamese task-oriented task-oriented dataset. 
The dataset contains 5,000 dialogues and 60,946 utterances, all of which are fully annotated on both the system and user sides.
We also do extensive experiments on a number of settings, from zero shot through mixed-language training on a variety of
architects and pre-trained language models. Our findings show that doing task-oriented dialogue system 
in a low-resource language context is still difficult, especially end-to-end manner.
The findings and analysis further show that, on low-resource scenario, while multilingual and bilingual LMs cannot achieve similar performance to that of models trained on high-resource language, multilinguality still has a positive impact on model's outcomes when compared to monolingual models, especially when pre-trained LMs are not available.

\textbf{Limitations and future work}. The fundamental limitation of our work is that the quantity of dialogues is still lower than in English, but this also corresponds to a real-world problem in which low-resource language data is not easily accessible and it is costly to annotate. In the future, we intend to expand the dataset by including side information such as co-reference resolution data, as well as conduct more experiments using machine translation algorithms for zero-shot and few-shot scenarios. Despite these limitations, we believe the ViWOZ is an essential step in the development of multilingual task-oriented dialogue systems.

\bibliography{anthology,custom}
\bibliographystyle{acl_natbib}

\appendix



\end{document}